\documentclass[sigconf]{acmart}
\AtBeginDocument{%
  \providecommand\BibTeX{{%
    \normalfont B\kern-0.5em{\scshape i\kern-0.25em b}\kern-0.8em\TeX}}}

\usepackage[utf8]{inputenc} % allow utf-8 input
\usepackage[T1]{fontenc}    % use 8-bit T1 fonts
\usepackage{hyperref}       % hyperlinks
\usepackage{url}            % simple URL typesetting
\usepackage{booktabs}       % professional-quality tables
\usepackage{amsfonts}       % blackboard math symbols
\usepackage{nicefrac}       % compact symbols for 1/2, etc.
\usepackage{microtype}      % microtypography
\usepackage{xcolor}         % colors

\usepackage{tikz}
\usepackage{pgfplots}
\usepgfplotslibrary{statistics}
\usepackage{enumitem}

\usepackage{amsmath,amsthm,bm} % Math packages
\usepackage{natbib}
\usepackage{cancel} 
\usepackage{caption}
\usepackage{subcaption}
\usepackage{threeparttable}
\usepackage{bm}
\usepackage{multicol}
\usepackage{multirow}
\newcounter{relctr} %% <- counter for relations
\everydisplay\expandafter{\the\everydisplay\setcounter{relctr}{0}} %% <- reset every eq
\AtBeginDocument{} %% <- store original definition

\definecolor{mycolor1}{RGB}{246, 126, 0}
\definecolor{mycolor2}{RGB}{120, 81, 151}
\definecolor{mycolor3}{RGB}{84, 144, 196}
\definecolor{mycolor4}{RGB}{205, 35, 33}

\definecolor{myblue1}{RGB}{35, 110, 160}
\definecolor{myblue2}{RGB}{80, 165, 215}
\definecolor{myblue3}{RGB}{165, 215, 245}

\pgfplotsset{linestyle/.style={%
        font=\rmfamily\Labelsize,
        width=0.25\textwidth,
        height=4cm,
        mark size=1.1pt,
        ylabel near ticks,
        xlabel near ticks,
        label style = {font=\small},
        tick label style = {font=\small, yshift=0.5ex},
        ylabel shift = -5 pt, 
        xlabel shift = -2 pt,
        title style={yshift=-1.2ex,font=\footnotesize},
        y tick label style={/pgf/number format/.cd,fixed,fixed zerofill,precision=3,/tikz/.cd},
        legend image post style={scale=0.5},
        every axis plot/.append style={semithick},
        legend style={font=\scriptsize, mark size=4pt},
        legend columns=1, legend style={/tikz/column 2/.style={column sep=10pt}},
        mark options={scale=1.7}}}

\pgfplotsset{barstyle/.style={%
    ybar,
    ylabel near ticks,
    xlabel near ticks,
    height=3.5cm,
    width=0.30\textwidth,
    label style={font=\small},
    tick label style={font=\small, yshift=0.6ex},
    title style={yshift=-1.2ex,font=\footnotesize},
    ylabel shift = 5 pt, 
    xlabel shift = -2 pt,
    tickwidth         = 3pt,
    xtick=data,
    legend style={font=\tiny, mark size=1pt},
    legend columns=1, legend style={/tikz/column 1/.style={column sep=0.1pt}},
}}

\pgfplotsset{boxstyle/.style={
    boxplot/draw direction=y,
    ylabel near ticks,
    xlabel near ticks,
    height=3.5cm,
    width=0.30\textwidth,
    cycle list={{mycolor1},{mycolor3}},
    label style={font=\scriptsize},
    tick label style={font=\scriptsize, yshift=0.6ex},
    xtick={1,2},
    title style={yshift=-1.2ex,font=\footnotesize},
    ylabel shift = -5 pt, 
    xlabel shift = -2 pt,
}}

\pgfplotsset{xbarstyle/.style={%
    xbar,
    width=0.225\textwidth,
    label style={font=\small},
    tick label style={font=\small},
    axis x line*=bottom,
    axis y line*=none,
    height=3.7cm,
    title style={yshift=-0.5ex,font=\small},
    ylabel shift = -5 pt, 
    xlabel shift = -5 pt,
    ylabel near ticks,
    xlabel near ticks,
    tickwidth         = 3pt,
    ytick=data,
    bar width=1mm,
    xlabel style={font=\footnotesize}
}}

\usepackage{balance}

\usepackage{array}
\newcolumntype{P}[1]{>{\centering\arraybackslash}p{#1}}

\usepackage{algorithm}
\usepackage{algpseudocode}
\setcopyright{acmcopyright}
\copyrightyear{2022}
\acmYear{2022}
\acmDOI{XXXXXXX.XXXXXXX}

\acmConference[SIGKDD'22]{KDD CUP 2022}{August 14--18,
  2022}{Washington, DC.}
\acmPrice{15.00}
\acmISBN{978-1-4503-XXXX-X/18/06}

\begin{document}

\title{KDD CUP 2022 Wind Power Forecasting Team 88VIP Solution}

\author{Fangquan Lin}
\authornote{These authors contributed equally to the paper.}
\email{fangquan.linfq@alibaba-inc.com}
\affiliation{%
  \institution{Alibaba Group}
  \city{Hangzhou}
  \country{China}
}

\author{Wei Jiang}
\authornotemark[1]
\email{alice.jw@alibaba-inc.com}
\affiliation{%
  \institution{Alibaba Group}
  \city{Hangzhou}
  \country{China}
}
\author{Hanwei Zhang}
\authornotemark[1]
\email{hanwei.zhanghw@alibaba-inc.com}
\affiliation{%
  \institution{Alibaba Group}
  \city{Hangzhou}
  \country{China}
}

\author{Cheng Yang}
\authornotemark[1]
\email{charis.yangc@alibaba-inc.com}
\affiliation{%
  \institution{Alibaba Group}
  \city{Hangzhou}
  \country{China}
}

% \renewcommand{\shortauthors}{Lin, et al.}
\renewcommand{\shortauthors}{Lin, Jiang and Zhang}
\begin{abstract}
KDD CUP 2022 proposes a time-series forecasting task on spatial dynamic wind power dataset, in which the participants are required to predict the future generation given the historical context factors. The evaluation metrics contain RMSE and MAE. This paper describes the solution of Team 88VIP, which mainly comprises two types of models: a gradient boosting decision tree to memorize the basic data patterns and a recurrent neural network to capture the deep and latent probabilistic transitions. Ensembling these models contributes to tackle the fluctuation of wind power, and training submodels targets on the distinguished properties in heterogeneous timescales of forecasting, from minutes to days. In addition, feature engineering, imputation techniques and the design of offline evaluation are also described in details. The proposed solution achieves an overall online score of -45.213 in Phase 3.
\end{abstract}





\maketitle
\section{Introduction}
KDD Cup 2022 proposes a dynamic wind power forecasting competition %of a wind farm at various timescales, 
to encourage the development of data mining and machine learning techniques in the energy domain
\cite{zhou2022sdwpf}.
%given the historical dynamic context and the spatial distribution of wind turbines \cite{zhou2022sdwpf}. 
In this challenge, wind power data are sampled every 10 minutes for each of the 134 turbines of a wind farm. Essential features consist of external characteristics (\textit{e.g.} wind speed) and internal ones (\textit{e.g.} pitch angle), as well as the relative location of all wind turbines. Given this dataset, we aim to provide accurate predictions for the future wind generation of each turbine on various timescales.

In this paper, we present the solution of Team 88VIP. %for the sequential time-series forecasting task. 
In general, we ensemble two types of models: a gradient boosting decision tree to memorize basic data patterns \cite{friedman2001gbdt, guyon2017lightgbm}, and a recurrent neural network to capture deep and latent probabilistic transitions \cite{mikolov2010recurrent,chung2014gru}. Ensemble learning of these two types of models, which characterize different aspects of sequential fluctuation, contributes to improved robustness and better predictive performance \cite{dong2020ensemble}.

Specifically, compared to general time-series forecasting tasks, distinguished properties of wind power forecasting are observed: \textit{i) Spatial dynamics}: Predictions for each of the 134 turbines are required. The distributions of two turbines are not fully matched, but share a certain spatial correlation \cite{damousis2004spatial}. \textit{ii) Timescales variation}: 288-length predictions are required ahead of 48 hours, which cover both the short-term and long-term prediction scenarios \cite{WANG2021117766}. \textit{iii) Concept drift:} Due to the large time gap between the test set and the training set, the distribution discrepancy cannot be ignored \cite{lu2019drift}. 

To tackle these specific challenges, corresponding model components are included. First, the spatial distribution of the turbines is considered for clustering data and imputing missing values. Secondly, submodels training and continual training are proposed to handle the heterogeneous timescales, for tree-based and neural network-based methods, respectively. Third, the distribution drift is adjusted during the inference stage, by comparing the expectation of the predicted values and the average of the ground truth.
%Empirically, comprehensive online and offline experiments are conducted to guarantee the model effectiveness. 
Empirical results demonstrate the effectiveness of the proposed solution. In all, the proposed solution achieves an overall online score of \textbf{-45.214} in Phase 3.

% \begin{itemize}
%     \item fluctuation and abnormal values
%     \item forecasting timescales
%     \item prediction data drift
% \end{itemize}
\section{Method}
\subsection{Solution overview}
\begin{figure*}[h]
  \centering
  \includegraphics[width=0.95\linewidth]{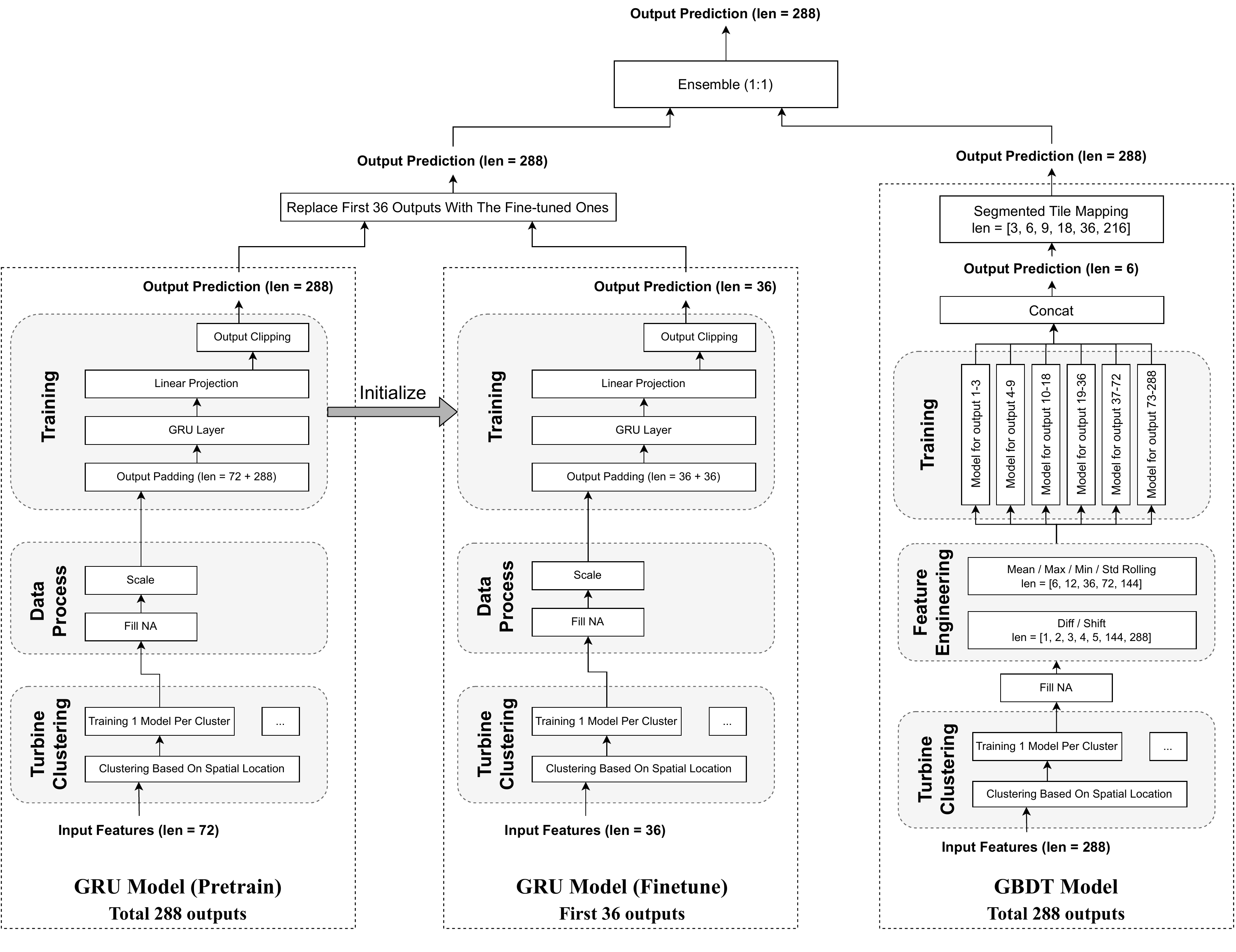}
  \caption{Total pipeline of the proposed solution.}
  \label{fig:pipeline}
\end{figure*}

In this section, we introduce the total pipeline of the proposed solution, as illustrated in Figure \ref{fig:pipeline}.
Four essential stages are involved, including: turbine clustering, data preprocessing, model training, and post-processing. And two major models are included through ensemble learning: Gate Recurrent Unit (\textbf{GRU}) and Gradient Boosting Decision Tree (\textbf{GBDT}). 

\subsection{Solution details}
\subsubsection{Major models}
\paragraph{Gate Recurrent Unit}
GRU is a type of recurrent neural network with a gating mechanism \cite{cho-etal-2014-properties}. In recent years, it has shown good performance in many sequential tasks, including signal processing and natural language processing \cite{10.1016/j.neucom.2019.04.044,athiwaratkun2017malware}. Through less active in the energy domain, there are studies shows that GRU is well-performed in smaller-sample datasets, compared to other variants of recurrent network \cite{chung2014empirical,gruber2020gru}. For formally, at time $t$, given an element $x_t$ in the input sequence, GRU layer updates the hidden state from $h_{t-1}$ to $h_t$ as follows:
\begin{equation*} \label{eq:gru}
\begin{split}
r_{t}= &\sigma\left(W_{i r} x_{t}+b_{i r}+W_{h r} h_{(t-1)}+b_{h r}\right)\,, \\
z_{t}= & \sigma\left(W_{i z} x_{t}+b_{i z}+W_{h z} h_{(t-1)}+b_{h z}\right)\,, \\
n_{t}= & \tanh \left(W_{i n} x_{t}+b_{i n}+r_{t} \circ \left(W_{h n} h_{(t-1)}+b_{h n}\right)\right)\,, \\
h_{t}= &\left(1-z_{t}\right) \circ n_{t}+z_{t} \circ h_{(t-1)}\,,
\end{split}
\end{equation*}
where $r_t$, $z_t$, and $n_t$ represent the reset, update and new gates, respectively; $\sigma$ is the sigmoid function and $\circ$ denote the Hadamard product.
\paragraph{Gradient boosting decision tree}
GBDT is an ensemble of weak learners, \textit{i.e.}, decision trees \cite{hastie2009elements}. Iteratively, boosting helps to increase the accuracy of the tree-based learner, to form a stronger model \cite{mason1999boosting}. In addition, both model performance and interpretability can be achieved \cite{wu2008top}.

\subsubsection{Main stages}
\paragraph{Turbine clustering}
As demonstrated in Figure \ref{fig:pipeline}, the first stage is clustering the 134 turbines. Data in each cluster can then be processed and trained individually to improve the efficiency and effectiveness of training. In terms of clustering method, neighbored turbines can be placed into the same cluster according to relative location. Alternatively, statistical correlation of wind power series can be also used. Empirically, we find that spatial clustering outperforms that relying on correlation, as further detailed in Section \ref{sec:exp-cluster}.
\paragraph{Data preprocessing}
In the second stage, we preprocess the dataset. To deal with the large amount of missing values and abnormal values, data imputation techniques are employed, either by the average within the cluster, or by a linear interpolation based on time series. For GRU, data scaler is applied on numeric features. Details are provided in Section \ref{sec:exp-pre}. For GBDT, feature engineering is performed to reconstruct the features used in the model training, including: the mean, max, min and standard deviation with the rolling in the historical sequences. More precisely, we consider that the lengths of the rolling window range from 6 to 144, \textit{i.e.}, from one hour to one day. In addition, values differences between the last timestamps and the history, ranging from last 10 minutes to two days, are also taken into account. 
\paragraph{Model training}
GRU and GBDT are trained individually for each turbine cluster. For both, the mean square error (MSE) is considered as the training loss. 

To deal with the heterogeneous timescales in the 288-length outputs, we separately develop training techniques for the two models. For GRU, we first pretrain the model with input length (=72) and output length (=288)  for multiple epochs (=20); after that, we finetune the solution with a relatively small input length (=36) and output length (=36), such that the finetuned solution focuses more on the short-term prediction. Finally, we replace the first 36 points in the pretrained outputs with the finetuned ones, to obtain the final prediction. And for GBDT, different submodels are trained for different timescales of outputs. In this way, each submodel can extract the important features separately for short-term prediction and long-term prediction. Detailed empirical studies are described in Section \ref{sec:exp-train}. 

\paragraph{Post-processing}
During inference, regular post-processing consists of clipping the predicted output to a reasonable range and smoothing the prediction curve to a certain level.
% \section{Method Highlights}
% \subsection{Utilizing spatial distribution of turbines} \label{sec:method-spatial}
% \subsubsection{Group training}
% \subsubsection{Spatial imputation}
% \subsection{Training for heterogeneous timescales}
% \subsubsection{Continual training}
% \subsubsection{Submodels training}
% \subsection{Overfitting Online Test Data}
Specifically, due to the large time gap
between the test set and training set, the distribution discrepancy
can not be ignored, especially in the online prediction task. In this case, adjusting the prediction results to the average of the ground truth contributes to an improvement of global predictive performance. A simple yet efficient way is to multiply each predicted value by a constant $\alpha > 0 $. Intuitively, if the average of all ground truth is greater than the average of all predictions, then the optimal $\alpha$ > 1; otherwise, $\alpha$ < 1. More formally, 
\begin{equation*}
\alpha = \arg \min \ell(\hat{y}, y; \alpha) = \sum_i{(\alpha \hat{y}_i - y_i)^2} + \sum_i{|\alpha \hat{y}_i - y_i|} \,,
\end{equation*}
where $\hat{y}$ denotes the predicted values and $y$ the ground truth. The existence of an optimal solution is guaranteed considering that the loss function $\ell$ is convex \textit{w.r.t.} $\alpha$.  Empirical experiments related to the adjustment parameters are described in Section \ref{sec:exp-alpha}.

\section{Experiments}
We conduct comprehensive experiments to demonstrate the effectiveness of our solution. In this section, we focus on \textit{i)} providing an overall predictive performance of both online and offline datasets; \textit{ii)} providing detailed ablation study to analyze the effectiveness of modules, mainly in an offline setting.
\subsection{Data preprocessing}\label{sec:exp-pre}
Firstly, we introduce how the official training and test data are preprocessed. 
\paragraph{Invalid cases imputation}
We encode the invalid conditions, including both the abnormal values and missing values, as ``NA'' (not available). Abnormal values include cases where the wind power is negative or equals zero while the wind speed is large \cite{zhou2022sdwpf}. As summarized in Table \ref{table:stats}, the invalid conditions account for about a third of the total samples. To handle these values, we perform a two-step imputation strategy to reduce uncertainty and bias in the prediction \cite{rubin1988overview, schafer1999multiple}. Firstly, we substitute the missing values with the average within the cluster; Secondly, for those not sharing any valid conditions within the cluster, we perform a linear interpolation based on time-series.
\begin{table}[h] %表格的浮动环境
 \centering
% \footnotesize
% \setlength{\tabcolsep}{4.2pt}
% \scriptsize
% \small
 \begin{threeparttable}
%  \captionsetup{font=small}
 \caption{Dataset statistics.}
%  \vspace{-0.3cm}
\begin{tabular}{lrrrr}
  \toprule %表头直线
  & \# day  & \# samples & \# abnormal & \# missing \\ %avg. user history length \\
\midrule
Training &  245 & 4\,727\,520 & 1\,354\,025 & 49\,518 \\
Test & 16 & 308\,736 & 104\,625 &  2\,663 \\
% Adressa-10weeks &  &  & 20M \\
% Plista & Unknown & 70\,353 & 1\,095\,323 & German \\
% Yahoo News & \\
\bottomrule %表底直线
\end{tabular}
  \label{table:stats}
 \end{threeparttable}
\end{table}

\begin{figure}[h]
  \centering
    \begin{subfigure}{0.235\textwidth}
  \includegraphics[width=1\linewidth]{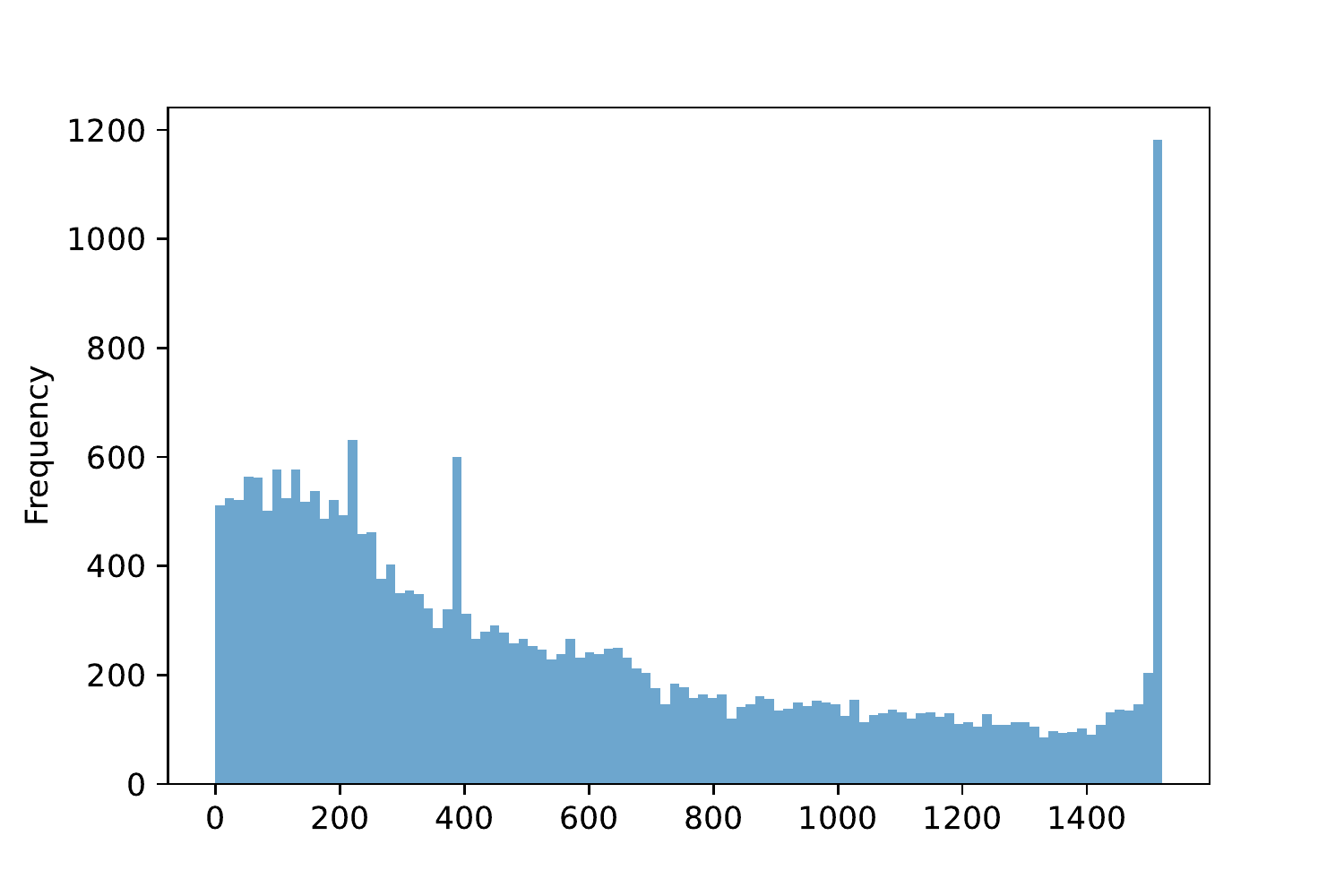}
  \caption{Wind power.}
  \end{subfigure}
  \hfill
   \begin{subfigure}{0.235\textwidth}
  \includegraphics[width=1\linewidth]{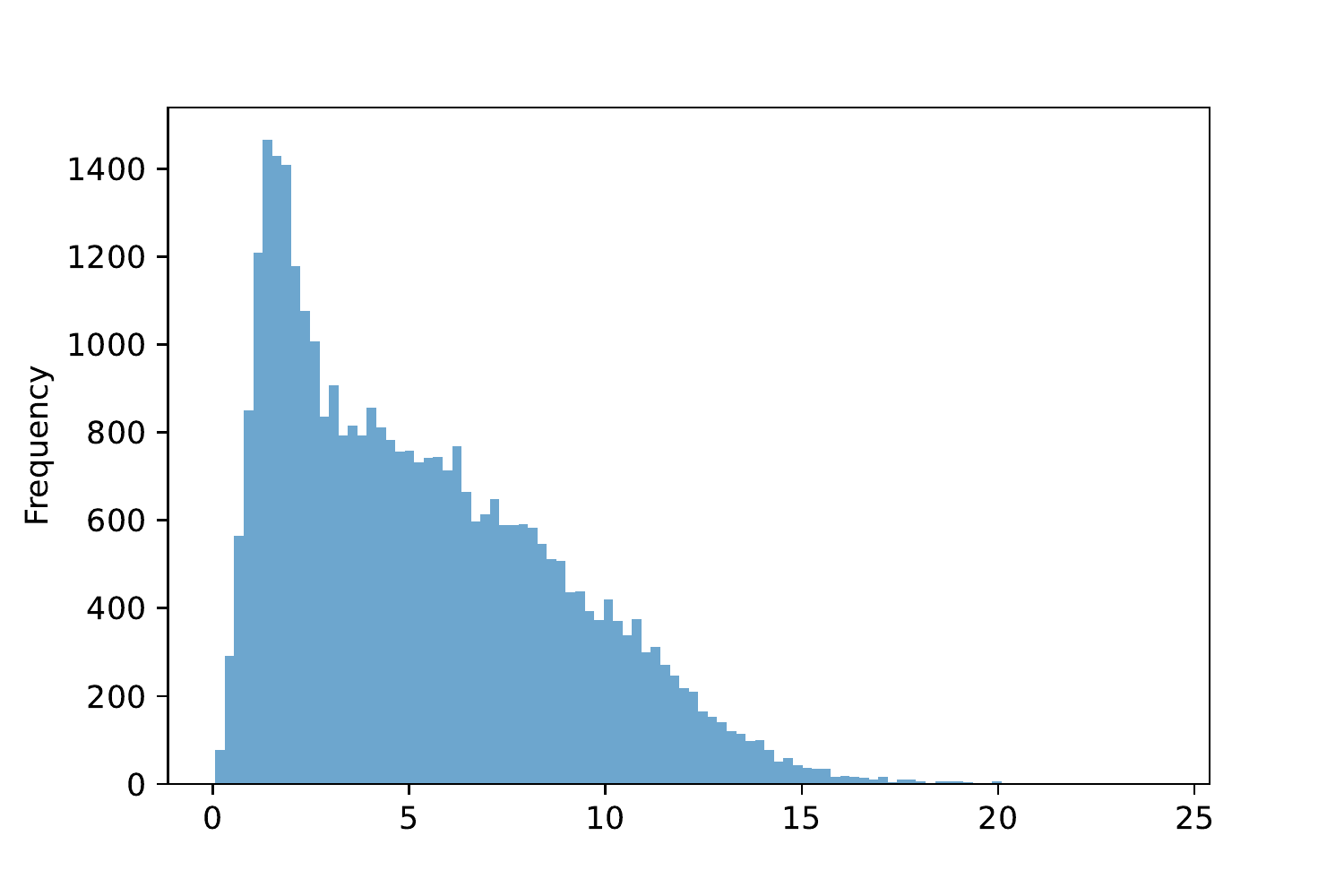} 
  \caption{Wind speed.}
  \end{subfigure}
  \caption{Visualizing data distribution. (\textbf{a}) Wind power. (\textbf{b}) Wind speed. Values less than 0 are removed in the plots.}
  \label{fig:data_distr}
\end{figure}

\paragraph{Data scaling} Following data imputation, we apply the feature scaler to keep the relative scales of features comparable. Considering that data %do not strictly follow the normal distributions 
have outliers as illustrated in Figure \ref{fig:data_distr}, a robust scaler is applied, which removes the median and scales the data according to the range between the $1^{\text{st}}$ quartile and the $3^{\text{rd}}$ quartile \cite{scikit-learn}. More formally, for each column $x$:
$$x_{\text{scaled}} = \frac{x - x_{\text{median}}}{x_{q75} - x_{q25}} \,.$$

\paragraph{Test set reconstruction}
The official test set defines a 2-day ahead forecasting evaluation given 14-day historical sequence. However, a single evaluation of one 288-length output is far from reliability. In addition, in our solution, the maximum input length is set as 288 (2 days). Thus we can reconstruct the test set, by randomly splitting the total 16 days to turn a single evaluation into 30 times through rolling windows. In this way, multiple evaluations guarantee the reliability of the offline results.
% In this way, we reconstruct the offline test set, to guarantee the reliability of the offline evaluation results.

% \textcolor{red}{
% \begin{itemize}
%     \item Build offline evaluation framework: splitting official test set
%     \item Statistics \& Distribution of wind power $\rightarrow$ Data scaler
%     \item Abnormal values $\rightarrow$ Imputation
% \end{itemize}
% }

\begin{table*}[t]
% \small
\centering
\caption{Online and offline forecasting performance of GBDT, GRU and the final ensemble models. The execution time is also provided. \textit{Improv} row shows the absolute improvements of the ensemble model over the best-performed single contributing model.}
\label{tab:performance}
% \ttabbox[0.96\FBwidth]{
%\vskip -0.1in
\begin{tabular}{lrrrrrrrrrr}
\toprule
 & \multicolumn{2}{c}{\textbf{Online}} & \multicolumn{6}{c}{\textbf{Offline}} &  \multicolumn{2}{c}{\textbf{Time}} \\
\cmidrule(lr{0.5em}){2-3} \cmidrule(lr{0.5em}){4-9} \cmidrule(lr{0.5em}){10-11} %\cmidrule(lr{0.5em}){8} 
 & Phase 3 & Phase2 &  Overall Score & RMSE & MAE & 6Hours Score & Day1 Score & Day2 Score & $\quad$ Train & Eval\\
\midrule
GBDT & - & -44.430 & -49.788 & 53.977 & 45.600 &  -32.537 & -46.526 & -53.032 & 40 min & 25 min\\
GRU  & - & -44.370 &  -49.797 & 53.924 & 45.670 & -32.946 & -46.434 & -53.051	& 100 min & 10 min\\
% \cmidrule(lr{0.5em}){2-6}\cmidrule(lr{0.5em}){7-11}
\midrule
Ensemble & \textbf{-45.213} & -44.195 & -49.646 & 53.775 & 45.517 & -32.406 & -46.302 & -52.935 & - & 35 min \\ 		
% \cmidrule(lr{0.5em}){2-6}\cmidrule(lr{0.5em}){7-11}
\textit{Improv}  & - & \textit{+0.175} & \textit{+0.142} & \textit{+0.149} & \textit{+0.083}  & \textit{+0.131} & \textit{+0.132} & \textit{+0.097} & - & -\\
% \textit{p-value} &  3.72\% & 0.25\% & 0.25\% & 0.25\% & 0.83\% & 2.34\% & 0.25\% & 0.25\% & 0.25\% & 0.25\% \\
\bottomrule
\end{tabular}
% \vskip -0.10in
\end{table*}

\begin{figure*}[t]
  \centering
  \includegraphics[width=1\linewidth]{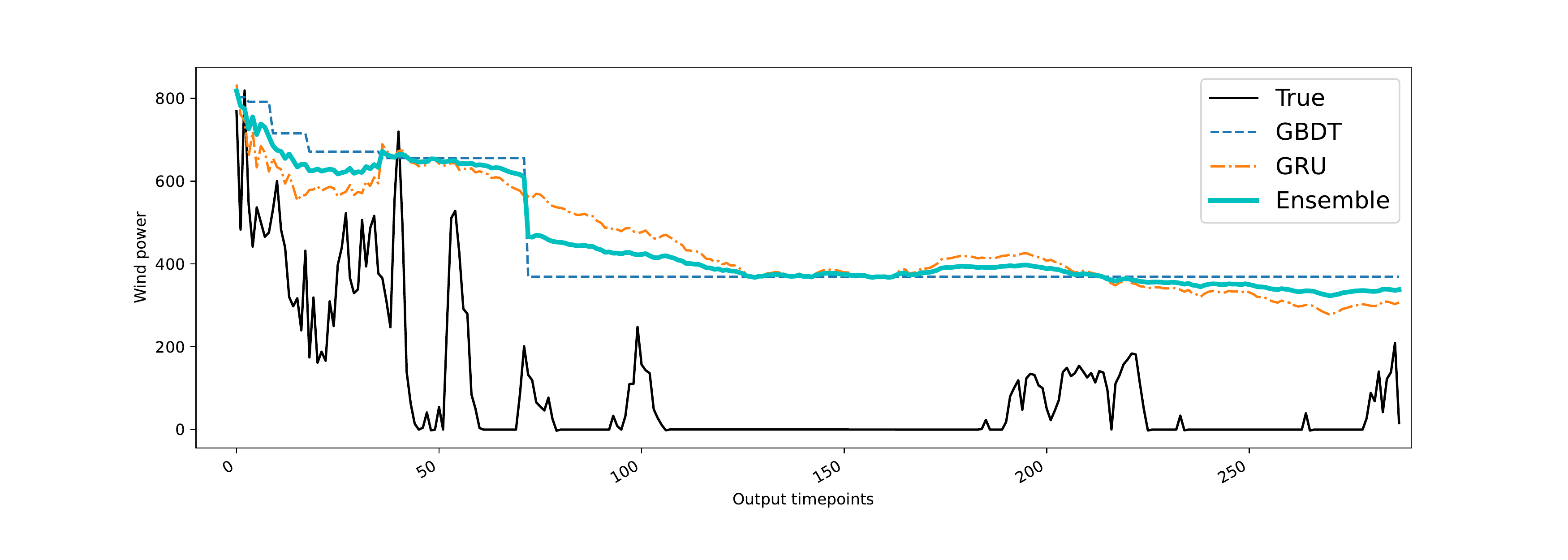}
  \vskip -0.30in
  \caption{An extract of 288-length prediction results.}
  \label{fig:plot_pred}
\end{figure*}

\subsection{Hyperparameter setting}
The hyperparameters for GRU are 
as listed in Table \ref{tab:hpgru}. Adam is applied for model optimization  \cite{DBLP:journals/corr/KingmaB14}.
For GBDT, we choose the first 200 days to train the model, while the remaining 45 days are used for hyperparameter tuning and early stopping. The maximum number of boost rounds is set as 1000 and the early stopping step is 20. Specifically, the other essential hyperparameters, including the number of leaves, the learning rate and the bagging parameters, are individually tuned for each submodel. Please refer to the codes for their precise values.

We conduct all the experiments on a machine equipped with a CPU: Intel(R) Xeon(R) Platinum 8163 CPU @ 2.50GHz, and a GPU: Nvidia Tesla v100 GPU.
% \textcolor{red}{
% \begin{itemize}
% \item summarize parameters in one or two tables
% \item Adam for GRU
% \item Machine information
% \end{itemize}
% }

\begin{table}[h] %表格的浮动环境
 \centering
% \footnotesize
% \setlength{\tabcolsep}{4.2pt}
% \scriptsize
% \small
 \begin{threeparttable}
%  \captionsetup{font=small}
 \caption{Main hyperparameter settings for GRU.}
%  \vspace{-0.3cm}
\begin{tabular}{lr}
  \toprule %表头直线
 Hyperparameter Names & Values \\ %avg. user history length \\
\midrule
GRU layers &    2 \\
GRU hidden units & 48 \\
Numeric embedding dimension & 42 \\
Time embedding dimension & 6 \\
ID embedding dimension & 6 \\
Dropout rate &  0.05 \\
Learning rate & $10^{-4}$\\
% LR scheduler & $\frac{1}{2}$ decay per epoch\\
% Adressa-10weeks &  &  & 20M \\
% Plista & Unknown & 70\,353 & 1\,095\,323 & German \\
% Yahoo News & \\
\bottomrule %表底直线
\end{tabular}
  \label{tab:hpgru}
 \end{threeparttable}
\end{table}

\subsection{Overall performance}
Table \ref{tab:performance} illustrates the overall performance of the investigated solution. The improvement of ensemble model over the best-performed single contributing model is also reported. For evaluation metrics, apart from RMSE, MAE, and the overall scores of the total 288-length prediction, we also consider the scores of various output timescales, including those of the first 6 hours (6Hours Score), the first day (Day1 Score) and the second day (Day2 Score). 

According to Table \ref{tab:performance}, the overall results are comparable between two single contributing models. However, in the offline setting, we can explore their slight variations among different timescales: On average, GBDT is better-performed in the first 6 hours prediction, while GRU surpasses GBDT in Day 1, which indicates the two models contribute diversely to the prediction. %and their performances in Day 2 are quite similar. 
Meanwhile, we can observe consistent improvements of the ensemble model on all metrics from offline to online. For instance, the ensemble model gains 14.20\% offline score, and 17.50\% online score (Phase 2) over the single contributing models. The improvements come from the reduction of prediction bias and variation by aggregating two type of models with diversity.

For a more intuitive comprehension, a representative example of forecasting results is showed as Figure \ref{fig:plot_pred}.
In addition, offline training and evaluation time are also reported in Table \ref{tab:performance}.

% \textcolor{red}{
% \begin{itemize}
% \item Compare ensemble score \& tree score \& GRU score;
% \item Training time
% \item 3 Prediction curves plot
% \end{itemize}
% }
\subsection{Effectiveness of model components}
Now we analyze how the various model components impact the forecasting performance. 

\subsubsection{Ablation study of turbine clustering}\label{sec:exp-cluster}
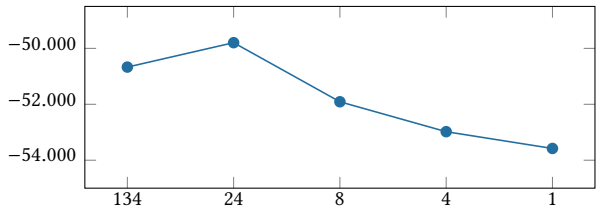
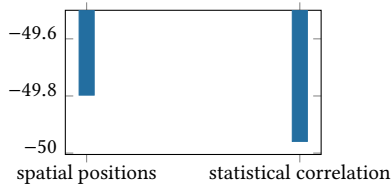
\begin{figure}[t]
    \centering
%     \begin{subfigure}{0.23\textwidth}
%       \begin{tikzpicture}
%             \begin{axis}[barstyle,width=\textwidth, title={}, xlabel={},bar width=2.9mm, symbolic x coords = {with, without}, legend style={draw=none, at={(0.4,0.8)}, anchor=west, nodes={scale=0.9, transform shape}, legend image post style={scale=0.5}}]
%             \addplot [color=myblue1, fill=myblue1] coordinates { 
%             (with, -49.797)
%             (without, -49.903)
%             };
%             \end{axis}
%             \end{tikzpicture}
%             \captionof{figure}{Spatial imputation.}%, evaluated in terms of GAUC and NDCG5 on MIND.}
% \end{subfigure}
    \begin{subfigure}{0.47\textwidth}
       \begin{tikzpicture}
            \begin{axis}[linestyle,width=\textwidth, title={}, xlabel={},bar width=2.9mm, symbolic x coords = {134, 24, 8, 4, 1}, legend style={draw=none, at={(0.4,0.8)}, anchor=west, ymax=-48.5, ymin=-55, nodes={scale=0.9, transform shape}, legend image post style={scale=0.5}}]
            \addplot [color=myblue1, mark=*] coordinates { 
            (134, -50.67)
            (24, -49.797)
            (8, -51.91)
            (4, -52.98)
            (1, -53.58)
            };
            \end{axis}
            \end{tikzpicture}
            % \vskip -0.05in
            \captionof{figure}{Number of turbine clusters (spatial positions). Specifically, Number 134 indicates the case of turbine-level individual modelling; Number 1 corresponds to a unified model for all turbines.}%, evaluated in terms of GAUC and NDCG5 on MIND.}
\end{subfigure}
 \vskip 0.15in
    \begin{subfigure}{0.28\textwidth}
       \begin{tikzpicture}
            \begin{axis}[barstyle,width=\textwidth, title={}, xlabel={},bar width=2mm, symbolic x coords = {spatial positions, statistical correlation}, legend style={draw=none, at={(0.4,0.8)}, anchor=west, ymax=-49.5, nodes={scale=0.6, transform shape}, legend image post style={scale=0.9}}]
            % \hskip 0.5cm
            \addplot [color=myblue1, fill=myblue1] coordinates { 
            (spatial positions, -49.797)
            (statistical correlation, -49.959)
            };
            \end{axis}
            \end{tikzpicture}
            % \vskip -0.05in
            \captionof{figure}{Clustering methods.}%, evaluated in terms of GAUC and NDCG5 on MIND.}
\end{subfigure}
\caption{Effectiveness of turbine clustering (GRU). (\textbf{a}) Comparison of the number of turbine clusters, according to spatial positions. (\textbf{b}) Comparison of clustering methods. }%utilizing spatial distribution (GRU).}
\label{fig:cluster}
\end{figure}

One of the most essential model components is clustering turbines to achieve a more effective and efficient training. More precisely, we compare the way of turbine clustering, either according to the spatial relative positions or to the statistical correlation of the wind power time-series. As demonstrated in Figure \ref{fig:cluster}, in practice, 24 clusters based on turbine location result in relatively good performance for GRU. Similarly, 4 clusters according to the turbine location is the best set-up for GBDT.

\begin{table}[h] %表格的浮动环境
 \centering
% \footnotesize
% \setlength{\tabcolsep}{4.2pt}
% \scriptsize
 \begin{threeparttable}
%  \captionsetup{font=small}
 \caption{Important features for submodels of GBDT.}
%  \small
%  \vspace{-0.3cm}
\begin{tabular}{ccc}
  \toprule %表头直线
Submodels &  Timescales& Top 3 important features \\ %avg. user history length \\
\midrule
\multirow{3}{*}{0 - 3} & \multirow{3}{*}{0-30 min (short-term)} &  Time \\
&  & Patv \\
& & Patv diff \\
\midrule
% 3 - 9 &  30 - 90min   &  time, Patv, Wspd\\
% 9 - 18 & 90min - 3h      &  time, Patv, Wspd \\
\multirow{3}{*}{18 - 36} &  \multirow{3}{*}{3-6h (mid-term)}         & Time \\
&  & Wspd \\
& & Patv mean rolling 72 \\
\midrule
% 36 - 72 &  6h - 12h        &  time, Wspd mean rolling 72, Patv mean rolling 144\\
\multirow{3}{*}{72 - 288} & \multirow{3}{*}{12h-2days (long-term)}     & Wspd max rolling 144 \\
&  & Wspd std rolling 144\\
& & Patv std rolling 144 \\
\bottomrule %表底直线
\end{tabular}
  \label{tab:importfeat}
 \end{threeparttable}
\end{table}

\subsubsection{Ablation study of heterogeneous timescales training}\label{sec:exp-train}
\begin{figure}[t]
 \centering
    \begin{subfigure}{0.43\textwidth}
       \begin{tikzpicture}
            \begin{axis}[barstyle,width=\textwidth, title={}, xlabel={},bar width=2.9mm, symbolic x coords = {7, 6, 2}, legend style={draw=none, at={(0.4,0.8)}, anchor=west, ymax=-49, ymin=-50.2, nodes={scale=0.9, transform shape}, legend image post style={scale=0.5}}]
            \addplot [color=myblue1, fill=myblue1] coordinates { 
            (7, -50.006)
            (6, -49.788)
            (2, -49.989)
            };
            \end{axis}
            \end{tikzpicture}
            % \vskip -0.05in
            \captionof{figure}{Number of submodels for GBDT. Number 6 corresponds to six submodels for output timescales $1-3-9-18-36-72-288$. Based on that, number 7 corresponds to splitting the last part to $72-144-288$. And number 2 for two submodels $1-72-288$.}
\end{subfigure}
 \vskip 0.1in
    \begin{subfigure}{0.28\textwidth}
       \begin{tikzpicture}
            \begin{axis}[barstyle, width=\textwidth, title={}, xlabel={},bar width=2mm, symbolic x coords = {with continual train, without continual train}, legend style={draw=none, at={(0.4,0.8)}, anchor=west, ymax=-49.5, nodes={scale=0.6, transform shape}, legend image post style={scale=0.9}}]
            % \hskip 0.5cm
            \addplot [color=myblue1, fill=myblue1] coordinates { 
            (with continual train, -49.797)
            (without continual train, -49.980)
            };
            \end{axis}
            \end{tikzpicture}
            % \vskip -0.05in
            \captionof{figure}{Continual training for GRU.}%, evaluated in terms of GAUC and NDCG5 on MIND.}
\end{subfigure}
\caption{Effectiveness of training on heterogeneous timescales. (\textbf{a}) Comparison of the number of submodels for GBDT. (\textbf{b}) Effectiveness of continual training for GRU. }%utilizing spatial distribution (GRU).}
\label{fig:timescales}
\end{figure}
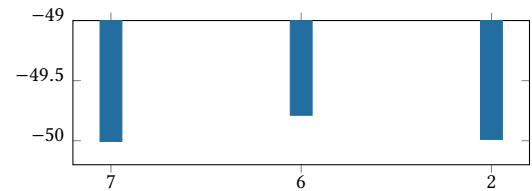
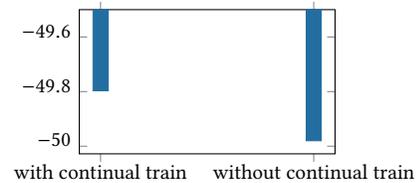

To deal with heterogeneous timescales in 288-length outputs, we train submodels for GBDT, and perform continual training for GRU, respectively. The effects of training techniques are shown in Figure \ref{fig:timescales}.

More specifically, we investigate the most important features for each GBDT submodel. Table \ref{tab:importfeat} shows that short-term and long-term predictions rely on different genres of features. For example, short-term predictions depend mostly on the latest values in the historical sequences, while long-term predictions focus more on the periodical statistical features such as the average wind power in the past 24 hours.

\subsubsection{Ablation study of the concept drift in prediction} \label{sec:exp-alpha}
% In the last 3 days of this competition, players have 9 chances to overfit the online test set of Phase 3. At this stage, we focus on exploring and incrementally correcting the predicted mean of our model to the true active power mean of the phrase3 test set. We multiply each predicted value by a global coefficient $\alpha$. If the true mean is less than the predicted mean, then $\alpha$ < 1, otherwise $\alpha$ > 1. After the model is fixed, the predicted value and the true value of the model are fixed, and the equation that minimizes the overall loss is convex. 

% $$
% Loss = \sum_i{(\alpha * pred_i - real_i)^2} + \sum_i{|\alpha * pred_i - real_i|} 
% $$

% In the actual process of the competition, we first tried $\alpha$ = 1.0, and the result was 45.45. Then tried $\alpha$ = 0.95 and got 45.75. Although the results have gotten worse, we have made it clear that the true value is larger than the predicted value. We tried two more times, and both online scores improved.

To correct the concept drift in prediction, the ablation study of adjustment parameter $\alpha$ is conducted in Phase 3 online, as shown in Table \ref{tab:alpha}.

\begin{table}[h] %表格的浮动环境
 \centering
% \footnotesize
% \setlength{\tabcolsep}{4.2pt}
% \scriptsize
% \small
 \begin{threeparttable}
%  \captionsetup{font=small}
 \caption{Effectiveness of adjustment parameter for online inference. \textit{Improvement} column indicates the absolute improvement \textit{w.r.t.} the score without adjustment ($\alpha=1$).}
%  \vspace{-0.3cm}
\begin{tabular}{ccc}
  \toprule %表头直线
Adjustment parameter $\alpha$ & Online Score (Phase 3) & \textit{Improvement}\\ 
\midrule
0.95 &    45.750  & \textit{-0.300}\\
1 (no adjustment) & 45.450  &  - \\
1.05 & 45.274 & \textit{+0.176} \\
1.10 & 45.213 & \textit{+0.237} \\
\bottomrule %表底直线
\end{tabular}
  \label{tab:alpha}
 \end{threeparttable}
\end{table}

Note that the ablation studies in one component have been conducted with all other hyperparameters equal. Through testing all possible combinations offline, consequently, the optimal solution is summarized in Table \ref{tab:performance}. Finally, for the reproducibility issue, we have repeated multiple times in the offline scenario, and the variation of the evaluation results is strictly controlled under 0.2\%.

\section{Conclusion}
In summary, GBDT memorizes the basic data patterns and GRU captures the deep and latent sequential transitions. Ensemble learning of these two models, which characterize different aspects of sequential fluctuation, leads to enhanced model robustness. The detailed design of each stage, including turbine clustering, data preprocessing, model training, and post processing, simultaneously contribute to a better predictive performance, empirically verified from offline to online.

There are several possible directions to explore in future work. First, more ensemble strategies can be considered, in addition to the average of prediction results. Secondly, farm-level forecasting task instead of the current turbine-level maybe interesting and beneficial in practice. Last but not least, uncertainty forecasting such as quantile regression or other probabilistic models is another center of new energy AI research that worth exploring.

% \section*{Acknowledgement}
% We thank Professor Wotao Yin who provided insight and support. 
% We thank Zhongkai Yi and Jiayu Han for comments and suggestions in the energy domain.
% We would also like to show our gratitude to Mengyang Niu and Wei Wang, who greatly assisted us.
% And we are immensely grateful to the KDD CUP Organizers for having held this event.
\bibliographystyle{ACM-Reference-Format}
\bibliography{sample-base}

% \appendix



\end{document}